\documentclass[conference10pt]{IEEEtran}

\usepackage[T1]{fontenc}
\usepackage[utf8]{inputenc}

\usepackage{siunitx}
\usepackage{graphicx}
\usepackage[backend=biber,maxnames=4,sorting=none]{biblatex}
\usepackage[shortcuts]{extdash}
\usepackage{ifthen}
\usepackage[caption=false]{subfig}
\usepackage{multirow}
\usepackage{array}
\usepackage{algorithm}
\usepackage{algpseudocode}

\graphicspath{{pics/}}
\newcommand{\defaultskip}{\medskip}
\newcolumntype{C}[1]{>{\centering}m{#1}}

\bibliography{xrad_paper}

\begin{document}

\title{Towards Dynamic Fault Tolerance for Hardware-Implemented
  Artificial Neural Networks: A Deep Learning Approach }

\author{
  \IEEEauthorblockN{ Daniel Gregorek, Nils Hülsmeier, Steffen Paul }\\
  \IEEEauthorblockA{
    Institute of Electrodynamics and Microelectronics (ITEM.me)\\
    University of Bremen, Bremen, Germany\\ % Institute for Theoretical Electrical Engineering and Microelectronics\\
    \{gregorek,huelsmeier,steffen.paul\}@uni-bremen.de\\
  }
}

\maketitle

\begin{abstract}
The functionality of electronic circuits can be seriously impaired by the occurrence of dynamic hardware faults. Particularly, for digital ultra low-power systems, a reduced safety margin can increase the probability of dynamic failures. This work investigates a deep learning approach to mitigate dynamic fault impact for artificial neural networks. As a theoretic use case, image compression by means of a deep autoencoder is considered. The evaluation shows a linear dependency of the test loss to the fault injection rate during testing. If the number of training epochs is sufficiently large, our approach shows more than 2\% reduction of the test loss compared to a baseline network without the need of additional hardware. At the absence of faults during testing, our approach also decreases the test loss compared to reference networks.
\end{abstract}

\vspace*{1mm}


\section{Introduction}

Dynamic hardware faults present a major impediment for the efficient
implementation of digital circuits \cite{kim_inexact_2014}. They can
occur e.g. due to noise at reduced safety margins in ultra low-power
electronics, process variations or circuit aging.  The faults can
result into bit-flips at memories or registers, impairing the
correctness of an electronic circuit
\cite{dodd_basic_2003}. Nevertheless exists a multitude of
applications, which do not depend on a completely exact result of
computation in the data path. Prominently, computer vision or mobile
communication intrinsically have to deal with noisy data from the
input sensors. Accordingly, artificial neural networks (ANN) present a
promising solution for stochastic computation near to the sensor
{\cite{plastiras_edge_2018}}. Although the exact data is thereby not
needed or available, it requires a high level of robustness and energy
efficiency with respect to the stochastic computation.

\defaultskip

Further requirements have to be considered for space missions, where
radiation additionally can cause run-time defects. Traditional
countermeasures like triple majority redundancy (TMR), temporal
redundancy or package shielding have additional overhead with respect
to the timing properties or payload weight
{\cite{schmidt2017temporal}}. Moreover, upcoming space missions
require an increasing capability of autonomy, e.g. the TRIPLE/nanoAUV
initiative intends to explore subglacial lakes at extraterrestrial
moons like e.g. Jupiter's Europa, which are however highly affected by
radiation \cite{waldmann2020triple} \cite{dachwald2020key}. While
space operations are quite reluctant to stochastic approaches, there
is nevertheless a strong need for advancements at least for the
on-board scientific computation and the efficient processing of the
sensor data. The satellite mission BIRD was the one of the first using
ANN for processing infrared images of the Earth's surface
{\cite{halle2000thematic}}. Similar, there is an ongoing trend to
image processing on-board of satellites \cite{yuhaniz2005embedded}
\cite{gregorek2019fpga}.

\defaultskip

While there is empirical evidence about intrinsic fault tolerance of
ANN to single event upsets (SEU) {\cite{velazco1999evidences}},
previous work generally deals with static stuck-at failures or
additional hardware overhead {\cite{torres-huitzil_fault_2017}}. In
\cite{kausar2016artificial} authors claim: “[..a] major influence on
the reliability of neural networks is the uniform distribution of
information”, from which a redundancy based approach could be
derived. In \cite{johnson2017homeostatic}, authors present a repair
mechanism for spiking neural networks using additional tuning
circuits. In \cite{su2016superior} authors propose a fault
injection-based genetic algorithm to construct a fault-tolerant
ANN. Orthogonal to that, this work investigates a customized deep
learning approach for improved dynamic fault tolerance using the
unmodified topology of a baseline ANN. While the assumptions about
intrinsic error resiliency of ANNs typically deals with static errors
or additional hardware overhead, this work is therefore not restricted
to that assumptions. In particular our paper has the following
contributions:

\begin{itemize}
  \item The hypothesis that ANNs can extrinsically learn the
    capability of dynamic fault tolerance.
  \item A training approach to generically improve the fault tolerance
    of deep artificial neural networks.
  \item A theoretic use case, where we apply our fault tolerance
    learning approach to image compression by means of autoencoders.
\end{itemize}

As a potential outcome of our work, the benefits of a robust network,
at a reduced safety margin and a lower power consumption, could be
leveraged for near-sensor image processing. The remainder of this
paper is organized as follows: Sec. II describes the training approach
and the implemented models, Sec. III presents the results of our
experiments and finally Sec. IV concludes our paper.

\section{Approach and Implementation}
\label{sec:impl}

This section describes the training approach, the fault model and the
architecture of the test network. Tab. \ref{tab:xrad} shows our
general approach: During training and testing, the ANN processes input
images $X$ and generates an output $Y$ depending on the configuration
of the ANN. The baseline network is affected by dynamic faults during
testing/deployment only. However, in this work we present a particular
approach how to beneficially inject dynamic faults during the training
of an ANN.

\begin{table}[htb]
  \centering
  \caption{General Approach}
  \label{tab:xrad}
  \begin{tabular}{l||p{2.4cm}|p{2.4cm}}
              & Training & Testing/Deployment \\\hline\hline
    Baseline  & \parbox[c]{1em}{\includegraphics[width=.12\textwidth]{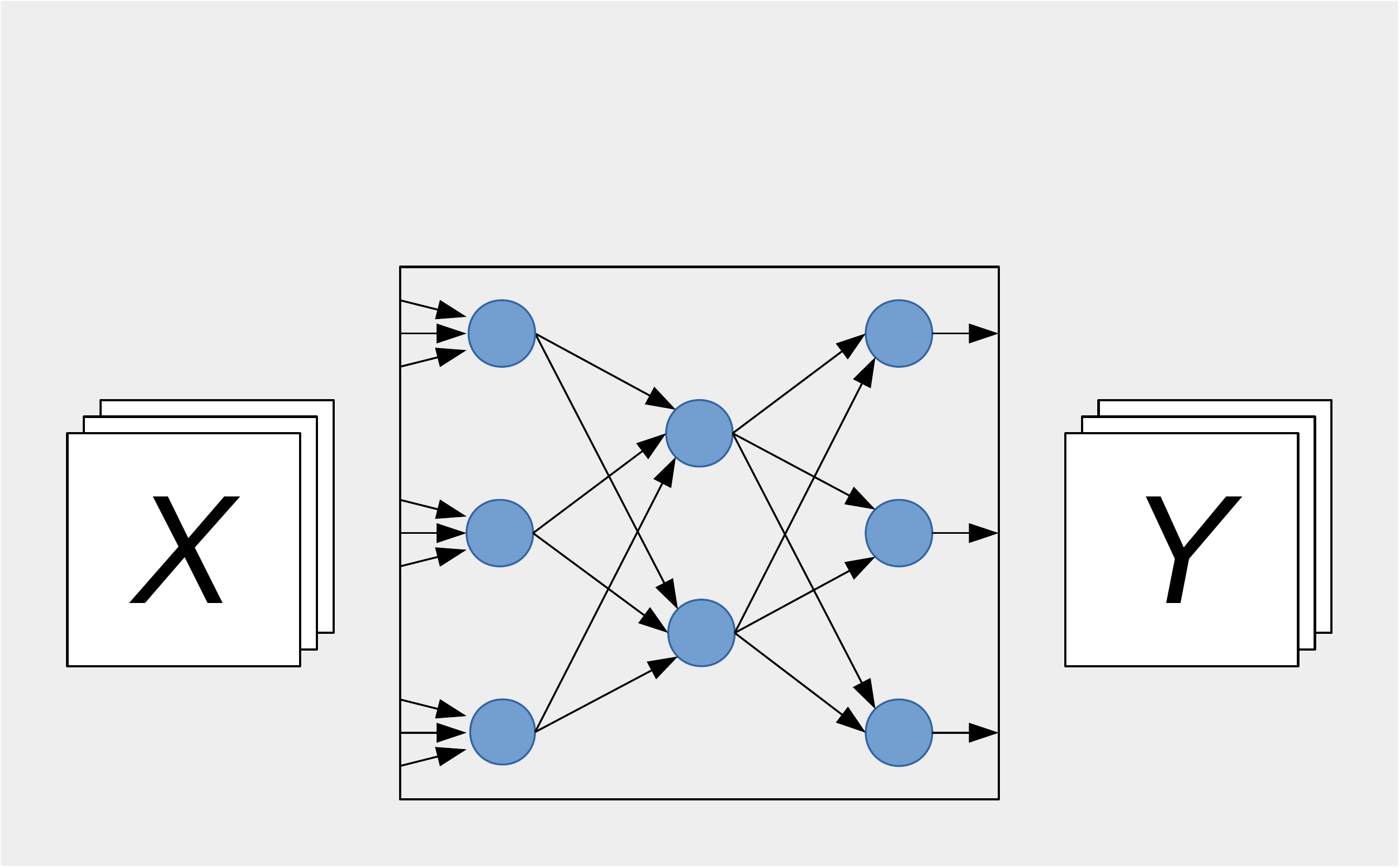}} & \parbox[c]{1em}{\includegraphics[width=.12\textwidth]{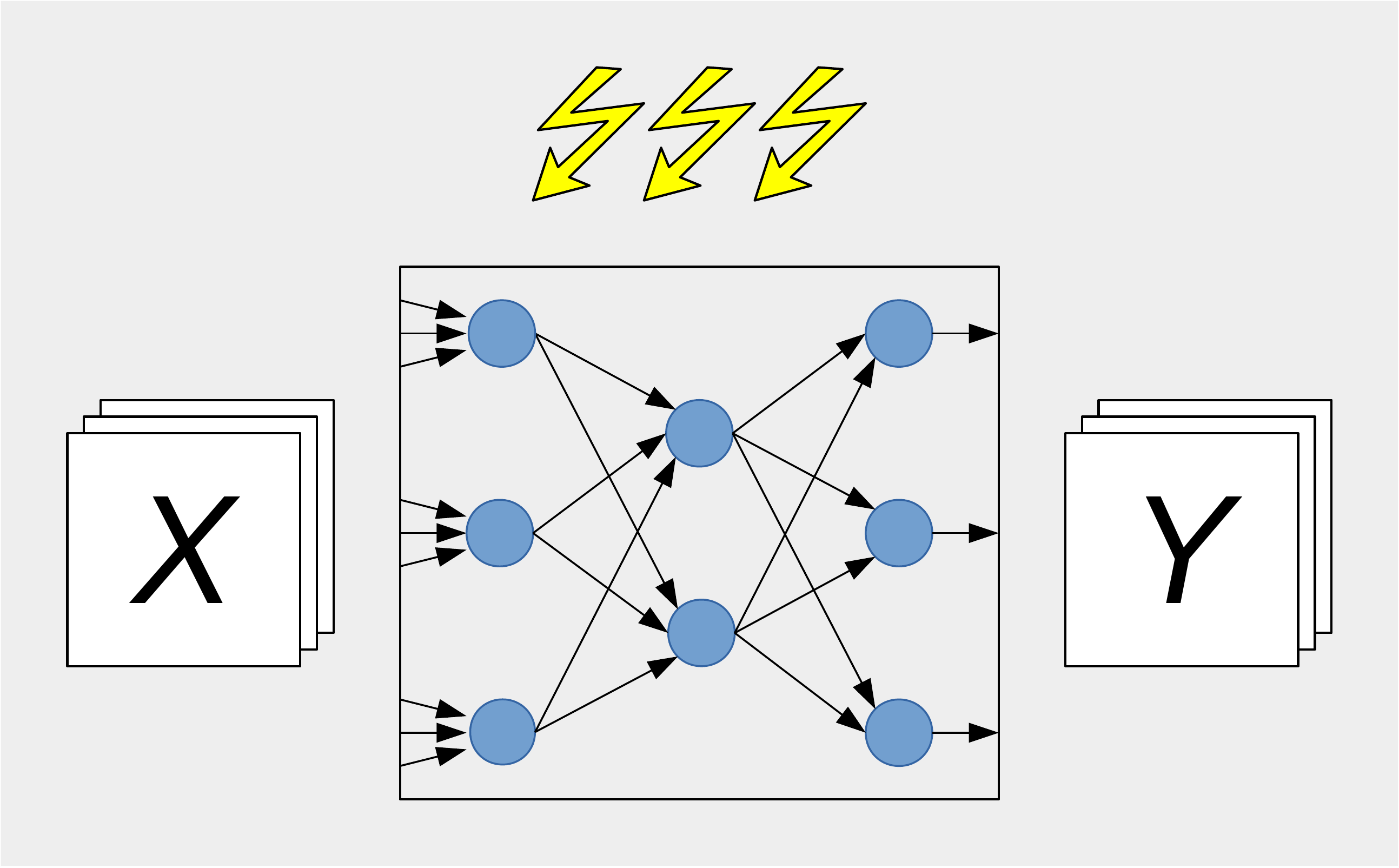}} \\\hline
    This work & \parbox[c]{1em}{\includegraphics[width=.12\textwidth]{xrad}} & \parbox[c]{1em}{\includegraphics[width=.12\textwidth]{xrad}} \\
  \end{tabular}
\end{table}

\subsection{Training Approach}

We propose a demonstration based approach to improve the fault
tolerance of ANNs by means of mock fault injection during ANN
training. The ANN is constituted by a directed acyclic graph $G$. The
faults are injected at the weights and biases of the ANN nodes
$N$. Each node $N_i$ corresponds to a particular layer $L_i$ of the
network graph $G$. Algorithm \ref{alg:mock} shows the general approach
how to perform a fault training epoch on a particular network graph
$G$ using the training data $(X_{train}, Y_{train})$. A random node of
the network is selected during training of each batch. A parameter of
the selected node is perturbed and the corresponding layer is frozen
while the remaining network is trained. We therefore call our approach
fault injection during training (FIT). Preceding to the fault training
epochs, one or more regular training epochs may be performed to adjust
the initial values of the parameters.

\begin{algorithm}
  \caption{Fault training approach}
  \label{alg:mock}
  \begin{algorithmic}[1]

    \Procedure {fault\_training\_epoch}{$G$, $X_{train}$, $Y_{train}$}
    \For{ each training-batch }
    \State random select node $N_i$ from $G$
    \State disturb node $N_i$ and freeze layer $L_i$
    \State perform regular training
    \State reset selected layer $L_i$
    \EndFor
    \EndProcedure
    
  \end{algorithmic}
\end{algorithm}

\subsection{Fault Model}

At the hardware level, the baseline ANN is assumed to be implemented
with signed fixed-point numbers. While the ANN input values are
typically normalized to range from -1 to +1, the internal values can
be of larger magnitudes inside the network. For practical reasons, we
assume a fixed bit-width of \mbox{$W=1+I+F$} bits for every value,
having one sign bit, $I$ integer bits and $F$ bits for the fractional
part.

\defaultskip

While for the training we inject exactly one fault per batch, a
variable amount of faults is injected during testing. To keep the
computational complexity (wall time) of the simulation short, we
generate a random number $x_i$ during evaluation of each test
batch. This random number $x_i$ is taken as the number of faults at
the network for that given batch. We choose a Poisson distribution to
to generate the number $x_i$ of independent random
faults. Subsequently, a random node is selected for each of the
faults. This approach avoids to iterate over all nodes, and simulate
whether a fault has occurred there.

\defaultskip

In our fixed-point model, any fault occurring at a node causes a
bit-flip in the corresponding fixed-point value. The affected bit
position is chosen by a uniform distribution. Additionally, to the
bit-flip fixed-point model, a tamed fault model using a Gaussian
distribution with zero mean, and configurable standard deviation is
also used.

\subsection{Test Architecture}

\begin{figure}[htb]
  \centering
  \includegraphics[width=.36\textwidth]{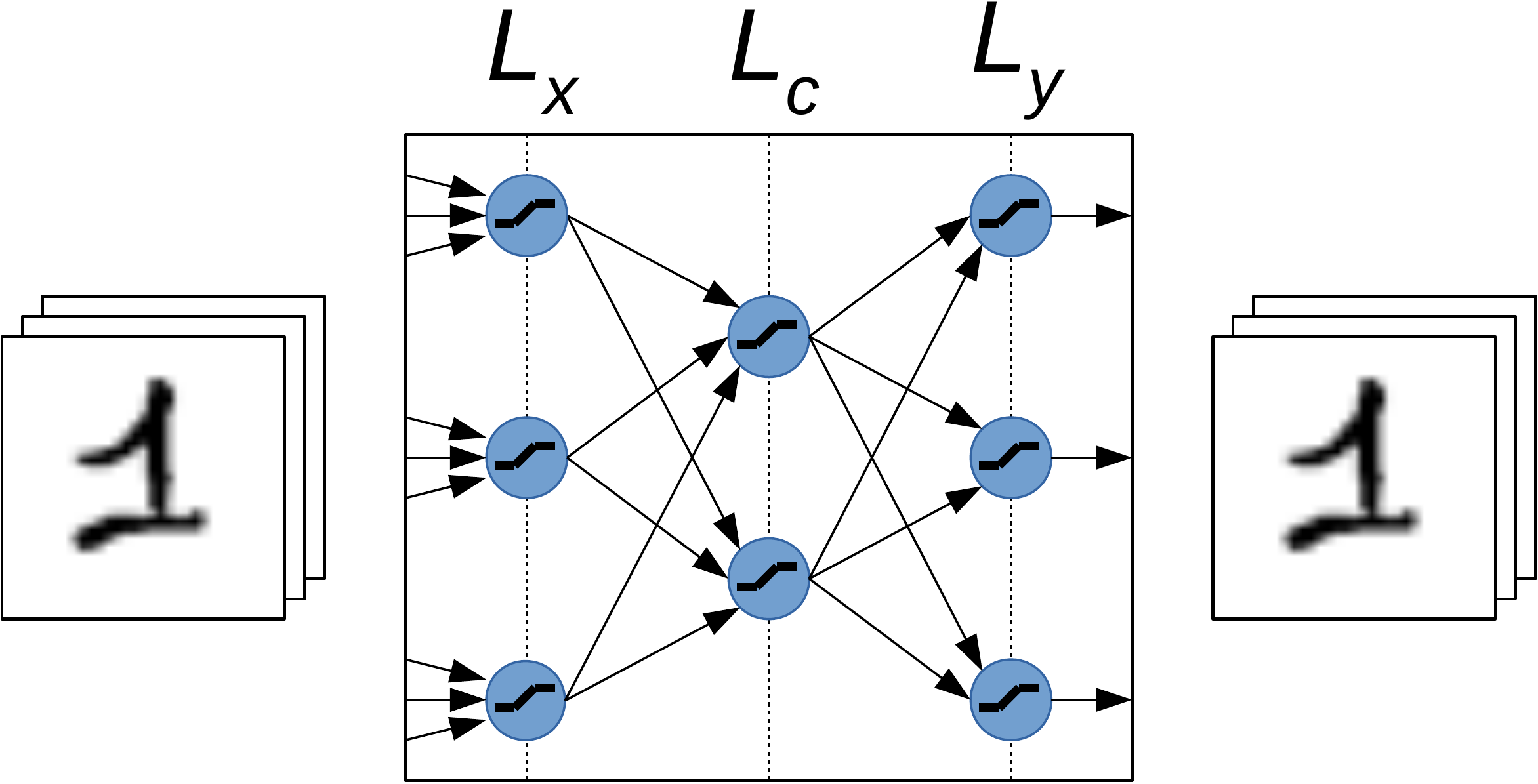}
  \caption{Sample autoencoder for image compression}
  \label{fig:ae}
\end{figure}

State-of-the-art convolutional and recurrent neural networks have a
complex structure potentially consisting of many specialized layers
and feedback paths. Due to its regular structure, we choose a deep
fully-connected autoencoder as a test architecture. Fig. \ref{fig:ae}
shows a sample architecture of the autoencoder (AE), which we use for
image compression. The given sample AE consists of an input layer
$L_x$, the center layer $L_c$ and an output layer $L_y$. The center
layer thereby determines the number of features, respectively the
ratio of the image compression. During training, the input images are
taken as the reference for the desired output of the AE, i.e. after
training the output $Y$ should match the input $X$.

\begin{figure*}[htb]
  \centering

  \subfloat[Golden Reference: Mean test loss vs. number of features]{\includegraphics[width=.48\textwidth]{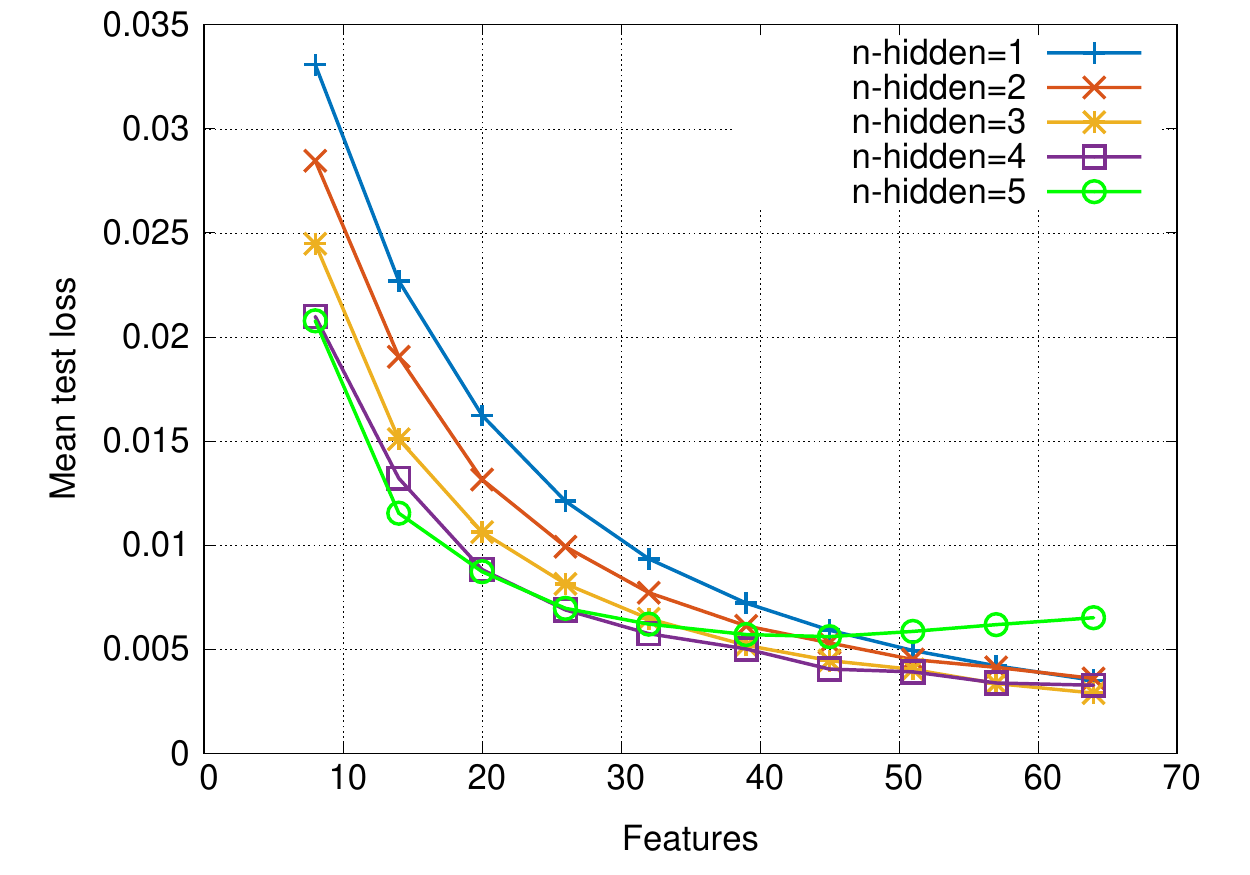}}
  \label{fig:eval:gold:feat}
  \subfloat[Golden Reference: Mean test loss vs. number of epochs]{\includegraphics[width=.48\textwidth]{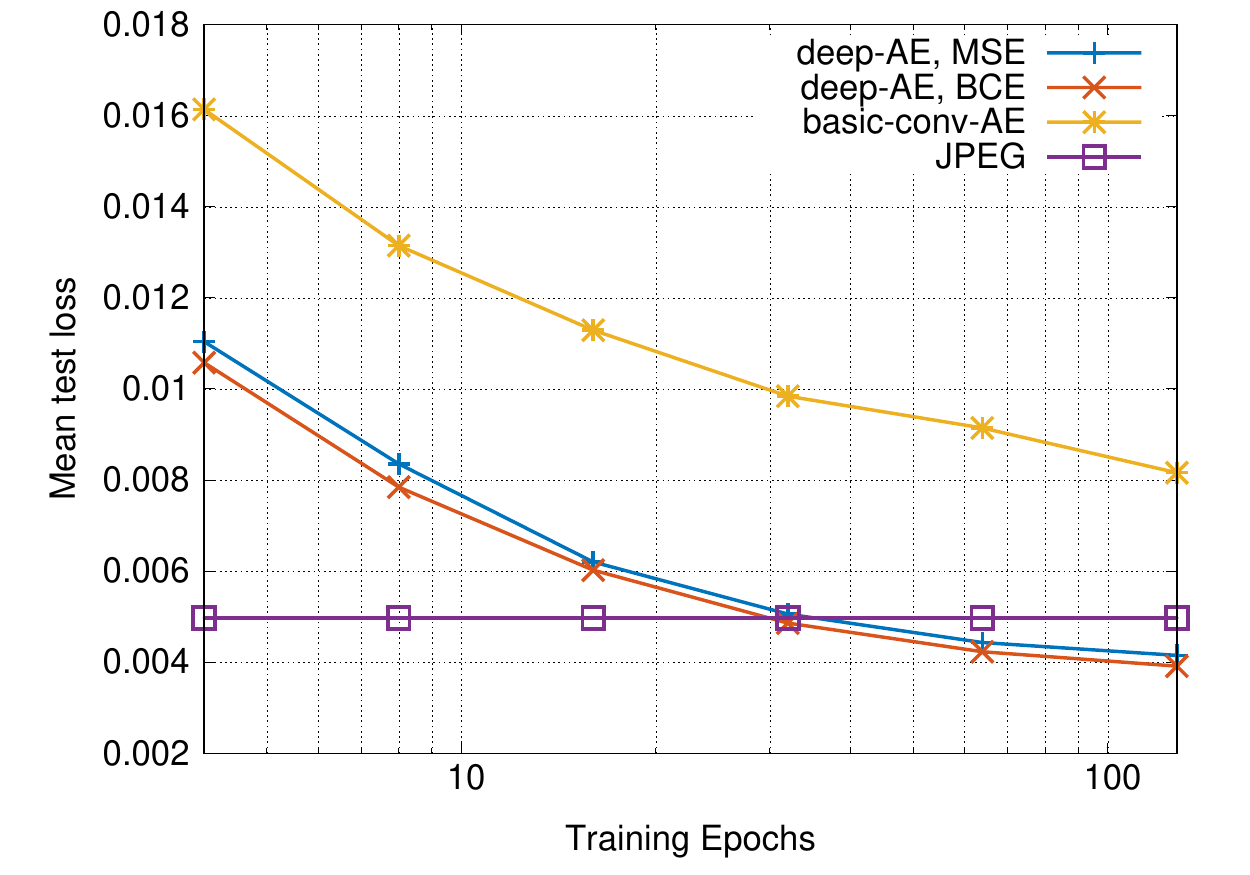}}
  \label{fig:eval:gold:jpeg}
  
  \vspace*{8mm}
  
  \subfloat[Mean test loss vs. fault rate]{\includegraphics[width=.48\textwidth]{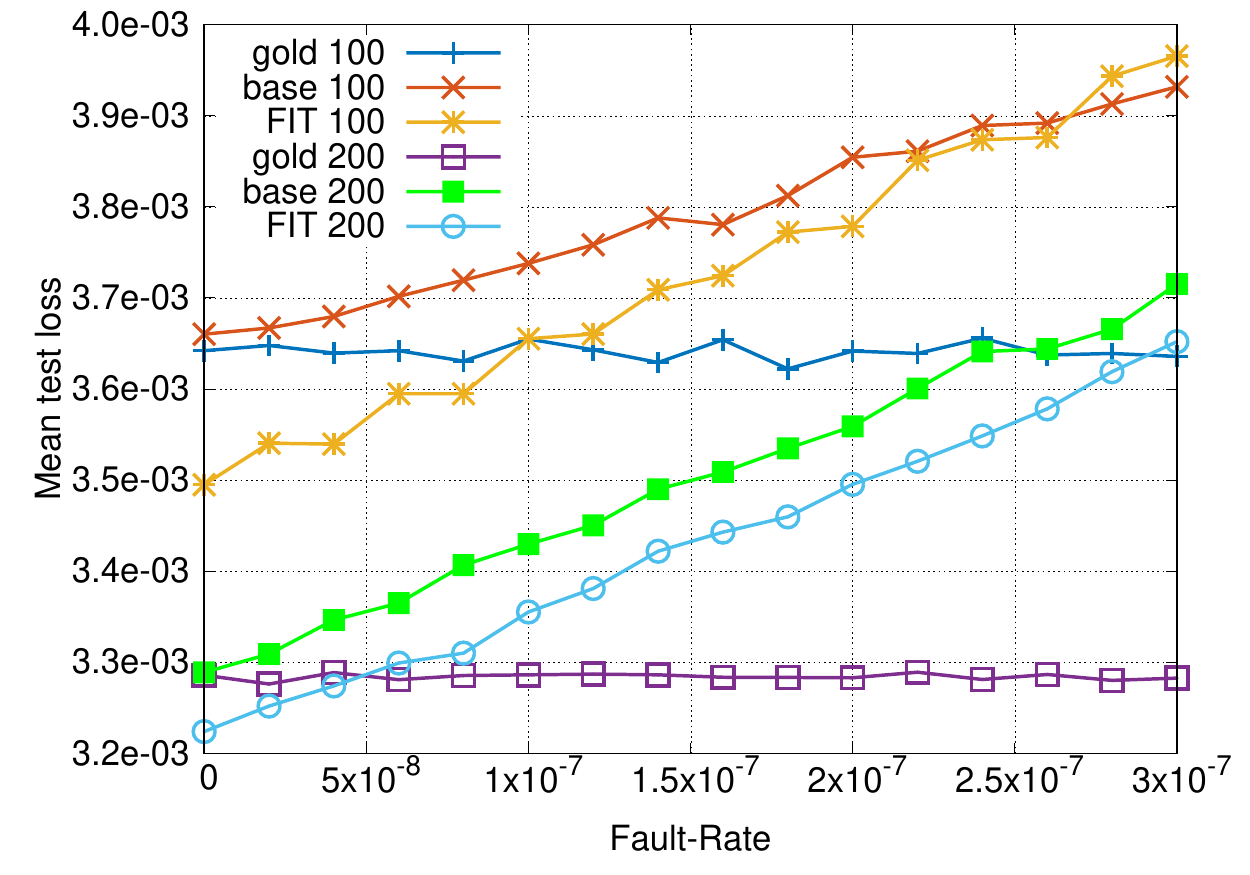}}
  \label{fig:eval:xrad:xpow}
  \subfloat[Mean test loss vs. number of training epochs]{\includegraphics[width=.48\textwidth]{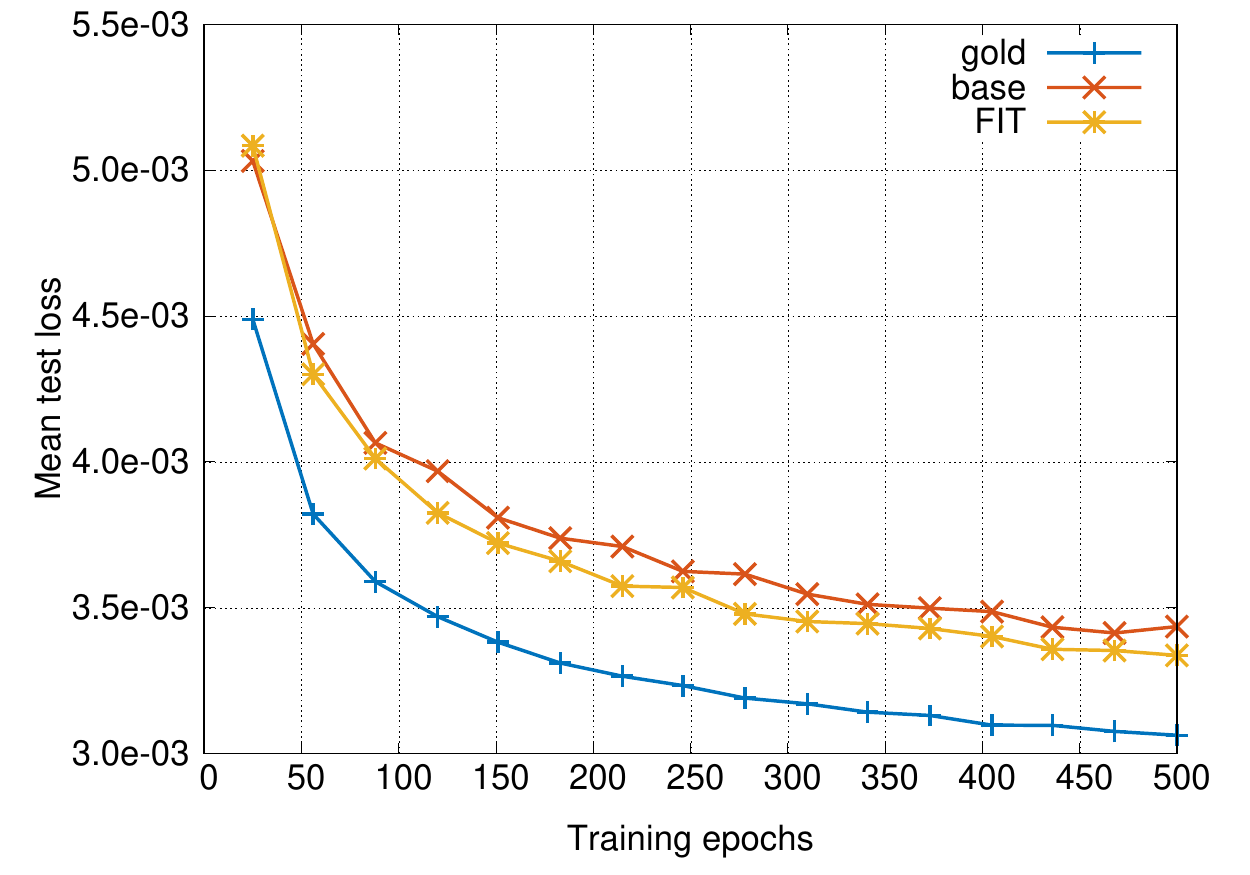}}
  \label{fig:eval:xrad:epoc}

  \caption{Experimental Results}
  
\end{figure*}

\section{Experimental Results}
\label{sec:eval}

The following section presents the experimental results with respect
to the image compression use case and our presented training
approach. We have used Tensorflow/Keras to implement the AE and the
fault injection using callback functions. Throughout the evaluation,
the MNIST dataset is used. Initially, an automatic hyperparameter
tuning using Ray Tune \cite{liaw_tune_2018} has been performed to find
reasonable starting values for the hyperparameters (e.g. batch size,
activation functions, etc.).

\subsection{Golden Reference}

To compare the results of our approach, a golden reference model has
been firstly developed. Fig.~\ref{fig:eval:gold:feat} shows the test
loss vs. the number of features / compression ratio of the golden
reference when using 100 training epochs. Further, the amount of
additional hidden layers is compared. Thereupon, an amount of 48 nodes
at the central layer and 3 hidden layers have been chosen to achieve a
reasonable trade-off between performance and complexity.

\defaultskip

Next, the performance of the AE is compared to the JPEG image
compression standard. Our input images are stored as 8bit TIFF
files. For the evaluation of JPEG, we convert the TIFF files to JPEG
and then back to TIFF. We compare the MSE between the true original
TIFFs vs. the reconstructed TIFFs. For the compression ratio, we
compare the number of bits for the input images versus the JPEG file
size. For the evaluation of the AE, we compare the MSE between the
input and the output images. The AE compression ratio is assumed as
the number of bits for the input image (28x28x8) vs. the number of
bits at the central layer (here 48x32). In our results, the AE
compresses the images to around 24.5\%, while JPEG achieves a
compression of approx. 33.3\%. Fig.~\ref{fig:eval:gold:jpeg} shows the
mean test loss of the AE in comparison to JPEG. After more than 30
training epochs, the AE outperforms JPEG regarding the MSE /
compression quality.

\subsection{Fault Effects}

Finally, the impact of the dynamic hardware faults and our proposed
training approach are evaluated. Due to the stochastic nature of the
hardware faults, our results are further averaged over multiple runs.
At training we have set a Gaussian fault model with $\sigma = 0.01$
and for testing the hardware related fixed-point fault model. For the
fixed-point format, a width W=16 bits and F=12 bits for the fractional
part have been set. Fig. \ref{fig:eval:xrad:xpow} shows the test loss
vs. the rate of injected faults during testing. We thereby define the
fault-rate as the average number of faults per node per sample, i.e. a
fault rate of \mbox{1e-7} roughly corresponds to 3.5 faults per 100
sample images. While the golden reference is not impaired by the
faults, the baseline using default training and our FIT approach using
mock fault injection during training have a linear dependency to the
fault rate. An exceptionally good result is shown for FIT at 100
training epochs and a small rate of injected errors. For 200 epochs
FIT generally has around 2\% smaller test loss compared to the
baseline. The improvement is quite small however it almost comes for
free since there is no need for additional hardware
overhead. Additionally, Fig. \ref{fig:eval:xrad:epoc} shows the test
loss vs. number of training epochs at a fault injection rate of 3e-7
indicating a robust improvement of the test loss for FIT after around
50 training epochs.

\section{Conclusion}
\label{sec:conc}

We presented a deep learning approach called fault injection training
(FIT) for artificial neural networks. The implemented fault model
allows the injection of hardware faults using the design framework
Tensorflow/Keras. The fault model uses bit-flips at fixed-point data
types or Gaussian weight perturbation. Our developed reference AE has
a superior performance for image compression compared to JPEG using
images from the MNIST dataset. The presented FIT training approach has
the general potential to reduce test loss and mitigates the impairment
due to hardware faults. Thereby, it does not require additional
hardware or software overhead during deployment of the neural network.

\defaultskip

This work presents an initial attempt to match the network training
more closely to the underlying stochastic properties of the hardware.
We plan a more in-depth analysis for the future. In particular the
detailed impact of the hardware faults needs to be better
understood. According to the results, a further refinement of the FIT
training approach should be possible. Further the required evaluation
wall time needs to be reduced. Due to the lengthy simulations,
e.g. results for a higher fault rate are still pending. Finally, the
approach needs to be evaluated for other deep neural network
architectures and use cases.

\printbibliography

\end{document}